\documentclass[journal]{IEEEtranTIE}
\usepackage{graphicx}
\usepackage{cite}
\usepackage{picinpar}
\usepackage{amsmath}
\usepackage{url}
\usepackage{flushend}
\usepackage[latin1]{inputenc}
\usepackage{colortbl}
\usepackage{soul}
\usepackage{pifont}
\usepackage{color}
\usepackage{alltt}
\usepackage[hidelinks]{hyperref}
\usepackage{enumerate}
\usepackage{siunitx}
\usepackage{epstopdf}
\usepackage{pbox}

\usepackage{upgreek}
\usepackage{mathtools}
\usepackage{booktabs}
\usepackage{multicol}
\usepackage{makecell}
\usepackage{multirow}
\usepackage[ruled, linesnumbered]{algorithm2e}
\usepackage[hidelinks]{hyperref}

\usepackage{changes}
\usepackage{lipsum}
\definechangesauthor[name={Per cusse}, color=orange]{per}

\def \textHT [#1]{\color{red}\textbf{#1}\color{black}}
\def \textLT [#1]{\color{blue}#1\color{black}}

\begin{document}
\title{3D Segmentation Learning from Sparse Annotations and Hierarchical Descriptors }

\author{
	\vskip 1em
	{
	Peng Yin, Lingyun Xu, Jianmin Ji, Sebastian Scherer and Howie Choset
	}

	\thanks{
		{
		This work was supported in part by the National Major Program for Technological Innovation 2030-New Generation Artificial Intelligence (No. 2018AAA0100500.)
		
        Peng Yin is with Robotics Institute, Carnegie Mellon University, Pittsburgh, PA 15213, USA (e-mail: pyin2@andrew.cmu.edu).
        Lingyun Xu is with SZ DJI Technology Co, Shenzhen, China (e-mail: hitmaxtom@gmail.com).
        Jianmin Ji is with School of Computer Science and Technology, University of Science Technology of China, Hefei, China (e-mail: jianmin@ustc.edu.cn).
		Corresponding author: Lingyun Xu (e-mail: xulingyun2021@gmail.com)
		}
	}
}

\maketitle
	
\begin{abstract}
    One of the main obstacles to 3D semantic segmentation is the significant amount of endeavor required to generate expensive point-wise annotations for fully supervised training. 
    To alleviate manual efforts, we propose GIDSeg, a novel approach that can simultaneously learn segmentation from sparse annotations via reasoning global-regional structures and individual-vicinal properties. 
    GIDSeg depicts global- and individual- relation via a dynamic edge convolution network coupled with a kernelized identity descriptor. 
    The ensemble effects are obtained by endowing a fine-grained receptive field to a low-resolution voxelized map. 
    In our GIDSeg, an adversarial learning module is also designed to further enhance the conditional constraint of identity descriptors within the joint feature distribution. 
    Despite the apparent simplicity, our proposed approach achieves superior performance over state-of-the-art for inferencing 3D dense segmentation with only sparse annotations. 
    Particularly, with $5\%$ annotations of raw data, GIDSeg outperforms other 3D segmentation methods.
\end{abstract}



\definecolor{limegreen}{rgb}{0.2, 0.8, 0.2}
\definecolor{forestgreen}{rgb}{0.13, 0.55, 0.13}
\definecolor{greenhtml}{rgb}{0.0, 0.5, 0.0}

\section{Introduction}

\IEEEPARstart{I}{n} autonomous driving, robotics and virtual reality, we usually obtain abundant 3D point cloud from ubiquitous sensing devices, such as LiDAR, RealSense and Kinect devices.
    The capability to directly measure point cloud is invaluable in those applications as 3D geometry could reduce segmentation ambiguities for scene understanding~\cite{Survey:Scene}, and 3D semantic information provides essential cues in decision making~\cite{Auto:2020survey}.
    While a number of 3D segmentation approaches have demonstrated promising results ~\cite{3D:pointnet,3D:pointnet2}, learning accurate point-wise segmentation requires large amounts of labeled training data~\cite{data:kitti,data:semantic3D}.
    Annotating 3D training data is a particular bottleneck in segmentation tasks, where labeling each point in the point cloud by hand is extremely time-consuming and requires expert knowledge in point cloud and complex 3D operation.
    This problem is illustrated on the \textit{SemanticKITTI} dataset~\cite{Data:semanticKitti} where finely annotating requires ``on average 1.5 hours for labeling a highway tile" and ``a total of over $1700$ hours" for the entire \textit{KITTI} odometry datasets~\cite{data:kitti}.
    
    \begin{figure}[t]
    	\centering
            \includegraphics[width=1\linewidth]{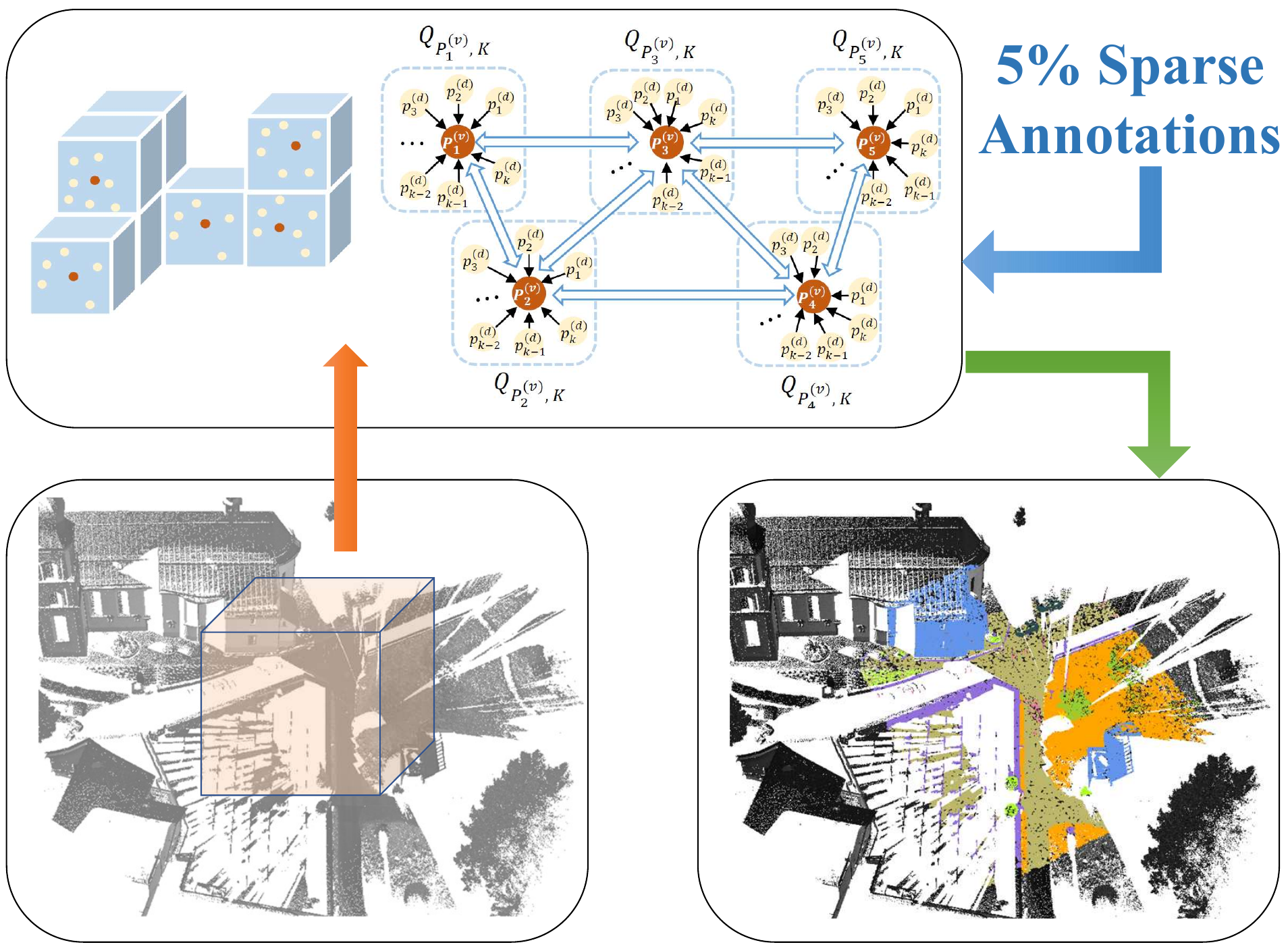}
    	\caption{\textbf{The proposed GIDSeg pipeline for 3D segmentation.}
        Given raw point cloud, we extract two scale (global-local) features for 3D semantic predictions.
        With only tiny 3D annotation labels, GIDSeg can achieve robust 3D segmentation.}
    	\label{fig:framework}
    \end{figure}

    In this paper, we focus on the problem of learning 3D segmentation using only sparse annotations which account for $5\%$ of the original 3D point cloud data. 
    However, such a task cannot be achieved by current point-based representations straightforward.
    Traditionally, volumetric representation of point cloud is a common approach~\cite{volumetric}, but it cannot capture high-resolution or fine-grained features even with excessive memory usage.
    PointNet-based methods~\cite{3D:pointnet,3D:pointnet2} treat points independently at local scale to achieve permutation invariance, but cannot capture geometric relationships among points, which reduces its inferencing ability for points with similar geometric structures.
    On the other hand, graph-based methods can infer points in both Euclidean and semantic space. For instance,  DGCNN~\cite{3D:DGCNN} constructs a graph and learns the embeddings for the edges on the global-local scale. However, it fails to capture geometric features on local scale.
    These methods either focus on independent points' features~\cite{3D:pointnet} on local scale, or target on extracting geometry structures~\cite{3D:DGCNN,volumetric} on global scale and directly ignore the local geometric structures.
    Thus the aforementioned works cannot capture plentiful features with only sparse annotations.

    To address these drawbacks, we propose a novel approach, called GIDSeg, which can achieve competitive 3D segmentation with only sparse annotations in the voxel level and reasoning concrete geometric structures from joint conditional features in the meantime.
    Instead of targeting on a single scale (i.e., sparse- or dense- scale ) feature extraction, GIDSeg generates global-local scale features that describe the relationships between global-local geometry structures via:
    (1) edge convolution operation~\cite{3D:DGCNN} on the global voxel points,
    and (2) radius kernelized feature extraction on the local points.
    The later works as local geometry features for voxel-wise global geometric features.
    The decoder of GIDSeg is designed to include both global and local geometric features, and thus the segmentation results are conditioned on the joint descriptions.
    Because the GIDSeg indirectly groups points in semantic space, the model is capable of inferencing unlabeled points' semantics based on labeled points with similar individual properties despite their distinct differences in Euclidean space.
    Using only sparse annotations corresponds to a reduction factor of $50$ in labeling \textit{KITTI} datasets~\cite{Data:semanticKitti}. 
    With such annotations, the performance of GIDSeg manifests its superiority over other point-based methods~\cite{3D:pointnet,3D:pointnet2,3D:pointconv} and fully annotated CNN-based approaches~\cite{3D:sqseg,3D:sqseg2,3D:rangenet++}. 
    Intrinsically, our work utilizes a cost-effective and sparse annotation-based strategy on 3D segmentation tasks.
	
    In our experiments, we compare our GIDSeg method with both point- and CNN- based state-of-the-art 3D segmentation methods, where GIDSeg and point-based methods are trained with sparse annotations (account $5\%$ in the original data) and CNN-based methods are trained with full annotations.
    On both \textit{Semantic3D} and \textit{SemanticKITTI} datasets, we show that the resulting network achieves competitive segmentation performance to all the point- and CNN- based methods with only sparse annotations. 
    In the ablation studies, we investigate the effect of different GIDSeg configurations 
    on the 3D segmentation performance.

\section{Related Work}
\label{sec:related_work}
    
    \subsection{Feature description in segmentation}
    Early approaches in point cloud segmentation can be divided into two categories: voxel~\cite{vol:splanet,vox:segcloud} and multi-view~\cite{3D:sqseg,3D:mvcnn} based methods. 
    Despite their considerable performance in public datasets~\cite{data:semantic3D,Data:semanticKitti}, they still have obvious shortcomings. 
    In most cases, 3D volume is sparse, and representing both occupied and free spaces as voxels makes it computationally intractable to perform CNNs on high resolution volumetric grids. 
    However, both multiview projections and voxelization will reduce the geometry details in original 3D point cloud.
    PointNet~\cite{3D:pointnet} is the first method that directly takes point sets as input, and explores the geometric interactions among neighboring points by integrating each point with a global signature. PointNet++~\cite{3D:pointnet2} further enhances the connections between local and global features by deploying PointNet in a hierarchical manner. 
    However, the points in local scale are treated independently for perturbation invariance, and their geometric connections are neglected.
    RandlaNet~\cite{3D:Randla-net} introduce a light-weight large-scale 3D segmentation method, which design a novel local feature aggregation method to increase the receptive filed for each 3D point.
    FusionNet~\cite{3D:DeepFusion} combine the point feature and voxel feature to achieve better 3D segmentation result.
    Zhu~\textit{et al.} include the Cylinder3D~\cite{3D:CylinderNet} via combining cylindrical partition and asymmetrical 3D convolution networks to explore the 3D geometric pattern.
    DGCNN~\cite{3D:DGCNN} introduces edge convolution that encodes the connections between points and proves that stacking multiple edge convolution layers can learn global shape properties. 
    But it fails to model the geometric feature of individual voxelized point.
    To alleviate this problem, radial basis function operators is applied~\cite{3D:PCNN} to represent geometric properties, and it shows invariance and equivariance to the raw data.

    To facilitate the robustness of local geometric representation while capturing the connections between them at the same time, we propose a hierarchical graph representation approach in our GIDSeg as described in Section~\ref{sec:HGE}. 

    \subsection{Learning from sparse annotation}
    \label{subsec:semi}
        Most fully supervised learning methods of 3D semantic segmentation require a large amount of point-wise labeled data, which are extremely costly to obtain. 
        Many trials on learning from sparse data are done on images. 
        Papandreou \textit{et al.}~\cite{papandreou2015weakly} introduced a method that learns from bounding box annotation of the objects. 
        But it can hardly be generated to point clouds, because in three-dimensional spaces, annotating each object bounding box is still non-trivial labor. 
        Qin \textit{et al.}~\cite{qin2018efficient} used sparse annotation to provide constraints for clustering. 
        However, it completely ignores the geometric relation between the annotated points. 
        FickleNet~\cite{lee2019ficklenet} generates a localization map from multiple combinations of random dropout on hidden layers to learn the relationship between locations in the image. 
        Wei \textit{et al.}~\cite{wei2018revisiting} introduces multiple dilated convolution layers with different dilation rates to produce a dense localization map. 
        However, generalizing these methods to point cloud can be expensive in both the number of random localization maps or dilation rate, and the convolution itself.
        3D Unet~\cite{cciccek20163d} is one of the few 3D semantic segmentation works with sparse annotations. 
        But their data is represented by voxel tiles of images and has an essential difference from the point cloud. 
        In the recent years, researchers also develop the 3D segmentation methods in the weakly supervised learning scope. 
        Wei~\textit{et al.}~\cite{3D:Multi_Path} introduce propose a weakly supervised approach to predict point-level results using weak labels on 3D point clouds.
        The most similiar work to ours is proposed by Xu~\textit{et al.}~\cite{3D:weakly}.
        The author designs a weakly supervised point cloud segmentation approach, which only needs a small labeled data in the training stage. 
        The major difference between our work and their work is that, our method introduce a hierarchical structure to capture both global spatial connections and local geometry structures.
        This property benefit in outdoor large-scale segmentation task as shown in the experiment.

    \begin{figure*}[t]
	    \centering
        \includegraphics[width=0.8\linewidth]{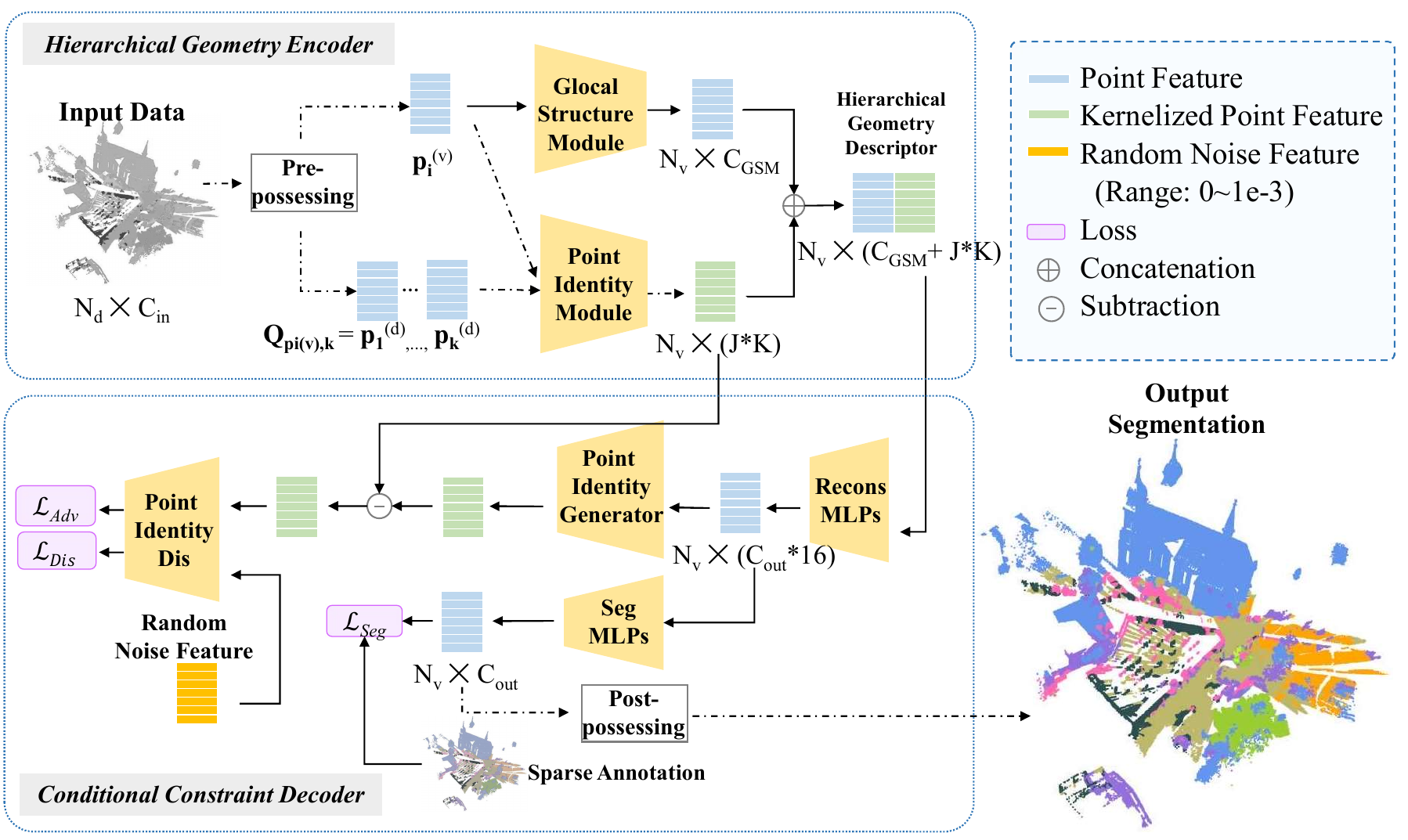}
	    \caption{\textbf{Architecture of GIDSeg for point cloud semantic segmentation}.The framework consists of a Hierarchical Geometry Encoder (HGE) for projecting the voxelized-level and dense level descriptors to a canonical space, coupled with a successive Conditional Constraint Decoder (CCD) to generate all labels for each points.} 
	    \label{fig:method}
    \end{figure*}
    
\section{Our Approach}
\label{sec:system}

    The proposed GIDSeg network operates on point cloud data with incomplete annotation
    to produce a full-resolution segmentation.
    As illustrated in Fig.~\ref{fig:framework},the framework of GIDSeg contains two major modules:(1) a Hierarchical Geometry Encoder (HGE) module, which encodes multi-scale feature distributions by combining the voxelized-level global features with a global-local Structure Module (GSM) and the dense-level features with a Point Identity Module (PIM); (2) a Conditional Constraint Decoder (CCD) module, which can enhance the connections between voxel- and dense- level during the segmentation predictions via a Generative Adversarial Networks (GAN) based adversarial learning procedure.
    The training labels are the voxelized-level points, which represent the incomplete sparse annotations and account for $5\%$ of raw data.
    
    
    To capture the spatial connections of points, we define a directed graph $\mathcal{G}=(\mathcal{V},\mathcal{E})$ representing the geometry structure, where $\mathcal{V}=\{v_1,...,v_n\}$ are the points, and $\mathcal{E} \subseteq \mathcal{V}\times \mathcal{V}$ are the edges between different points. Let superscript $d$ represent the dense raw points , $v$ represent the voxelized-level points. We denote dense points as $\mathbf P^{(d)} = \{\mathbf p_1^{(d)}$,...,$\mathbf p_{n_d}^{(d)}\}\subseteq\mathbf{R}^D $, voxelized points as $\mathbf P^{(v)} = \{\mathbf p_1^{(v)}$,...,$\mathbf p_{n_v}^{(v)}\}\subseteq\mathbf{R}^D $, where the dimension $D$ of the vertices node may vary based on the filter configurations. For the setting of $D$=4,  each point $\mathbf p_i=(x_i,y_i,z_i,i_i)$ is composed of 3D coordinates and LiDAR intensity. 
    The framework for HGE and CCD is described in Section~\ref{sec:HGE} and Section~\ref{subsec:CCD}.

    \subsection{Hierarchical Geometry Encoder}
    \label{sec:HGE}
    
        
        HGE enables integration of descriptors from both global-to-local and point-wise scope, providing a wealth of contextual information for point cloud semantic learning. Given a voxelized points $\mathbf p_i^{(v)}$, we formulate HGE to encode their geometry connections as follows:

        \begin{align}
            &E_{\Uptheta} (\mathbf p_i^{(v)}) = \nonumber \\ &E_{\Uptheta} [f_{GSM,\uppsi}(\mathbf p_i^{(v)},\mathbf p_j^{(v)}), f_{PIM}(\mathbf p_i^{(v)}, \mathbf Q_{\mathbf p_i^{(v)},K})]
        \end{align}
        where E represents HGE, $\{\mathbf p_j^{(v)}: (i,j)\subseteq\mathcal{E}\}$ serves as the neighbour patch of point $\mathbf p_i^{(v)}$, and $\mathbf Q_{\mathbf p_i^{(v)},K}=\{\mathbf p_1^{(d)},\mathbf p_2^{(d)},...,\mathbf p_K^{(d)}\}$ is the set of  the $K$-nearest points of $\mathbf p_i^{(v)}$ from the dense scope. Correspondingly, $f_{GSM,\uppsi}$ is the function of {global-local Structure Module} (GSM) that establishes global-regional construction of center point $\mathbf p_i^{(v)}$ and its neighbour information $\mathbf p_j^{(v)}-\mathbf p_i^{(v)}$, while $f_{PIM}$ represents the {Point Identity Module} (PIM) , providing dense-level local contextual information of point $\mathbf p_i^{(v)}$. $\Uptheta$ and $\uppsi$ are learnable parameters. Intuitively, $f_{GSM,\uppsi}$ shows how $\mathbf p_j^{(v)}$ relates to $\mathbf p_i^{(v)}$, learning the manner of grouping voxelized points within a point cloud, while $f_{PIM}$ is the point identity descriptor with abundant local region information incorporated, providing more details and higher accuracy.
        
        Given a dense point cloud, each voxelized point $\mathbf p_i^{(v)}$ and its corresponding $\mathbf Q_{\mathbf p_i^{(v)},K}$ can be obtained via a pre-possessing step. 
        We accumulate the 3D points into local dense map with the assistance of LiDAR odometry,
        then downscale the dense map into both lower resolution ($0.2m$) and higher resolution ($0.1m$) for point extraction from both GSM module and PIM module respectively.
        The concatenation of these two level descriptors are combined into Hierarchical Geometry Descriptor as shown in the Fig.~\ref{fig:method}.
        
 

    \subsubsection{Global-local Structure Module}
    \label{subsubsec:GSM}
       (GSM), denoted as $f_{GSM}$, is a combination of multiple EdgeConv layers~\cite{3D:DGCNN}, proven to efficiently capture the global-local feature connections without discarding points after applying farthest point sampling (FPS) like PointNet{\scalebox{.85}{++}}~\cite{3D:pointnet2} or HDGCNN~\cite{HDGCNN}. 
        Instead of connecting all the vertices in $\mathcal{V}$, we only build connections between each node and its $M$-nearest nodes in the graph $\mathcal{G}$. 
        The $M$-nearest nodes for the point  $\mathbf p_i^{(v)}$ are found by comparing the feature similarity in a radial distance function. 
        The output edge features between point $\mathbf p_i^{(v)}$ and $\mathbf p_j^{(v)}$ after multiple EdgeConv layers are defined as $f_{GSM,\uppsi}(\mathbf p_i^{(v)},\mathbf p_j^{(v)})$, where $f_{GSM,\uppsi}:\mathbf{R}^D \times \mathbf{R}^{D^{'}}$ is a nonlinear function with a set of learnable parameters $\uppsi$, which is define by
        \begin{align}
           f_{GSM,\uppsi}(\mathbf p_i^{(v)},\mathbf p_j^{(v)}) = \hat{f}_{GSM,\uppsi}(\mathbf p_i^{(v)},\mathbf p_j^{(v)}-\mathbf p_i^{(v)}).
        \end{align}
       
       The method explicitly combines global shape structure, captured by the coordinates of the patch centers $\mathbf p_i^{(v)}$, with its contextual information captured by $\mathbf p_j^{(v)}-\mathbf p_i^{(v)}$. Consider two layers $L$ and $L+1$ from adjacent EdgeConv's, the output of layer $L+1$ should be ${\mathbf p_j^{(v)}}^{'}=max_{j:(i,j)\in \mathcal{E}}h^{L}_{ij}$, where $h^{L}_{ij}=LeakyReLU[\upphi(\mathbf p_j^{(v)}-\mathbf p_i^{(v)})+\Omega\mathbf p_i^{(v)}]$. $\upphi$ and $\Omega$ are learnable parameters.
        Given a voxelized point cloud as input size $[N_v, C_{in}]$, we can obtain the feature from distance connection $\hat{f}_{GSM,\uppsi}(\mathbf p_i^{(v)},\mathbf p_j^{(v)}-\mathbf p_i^{(v)})$ in the vicinity with the output size of $[N_v, C_{GSM}]$. In our implementation, four Edge Conv's are used with output channels $EC_1=64, EC_2=64, EC_3=128$ and $EC_{4}=256$. The concatenation of the point feature outputs after each Edge Conv layers can be viewed as the global spatial connection descriptor with dimension $N_v\times C_{GSM}$.
            
    \subsubsection{Point Identity Module}
    \label{subsubsec:IM}
        (PIM) learns individual properties of points $\mathbf p_i^{(v)}$'s, where we use a combination of radius-based kernels to reflect the radial similarity. Let $\gamma$ be the inverse of the radius of influence of  $\mathbf Q_{\mathbf p_i^{(v)},K}$ on point $\mathbf p_i^{(v)}$, we consider $J$ different parameter $\gamma$ to set widths of the bell-shaped curve, expressing variety in influence of $\mathbf Q_{\mathbf p_i^{(v)},K}$.
        Then latent nearest-neighbor interpolation is applied where semantics are mapped in a radial manner. 
        Each point, as an identity, can be depicted in concatenation of $J$ radius-based kernel values: 
        \begin{align}
            &f_{PIM}(\mathbf p_i^{(v)}, \mathbf Q_{\mathbf p_i^{(v)},K}) \nonumber \\ 
            &= \odot_{j} exp[-\gamma_{j}\parallel \mathbf p_i^{(v)}-\mathbf Q_{\mathbf p_i^{(v)},K}\parallel^2], _{j=1,2,...,J}
            \label{rbf}
        \end{align}

        PIM intrinsically obtains the local geometry features for each voxelized identity $\mathbf p_i^{(v)} $ with the parameters $\gamma$ pretrained on dense maps, providing a auxiliary descriptor to HGE $K$-nearest neighbour results in $[N_v, J\times K]$ dimensional descriptor.
        Our best implementation takes $J=4$ and $K=16$.
        With this extracted hidden features, we are able to predict the semantic information with the multi-layer perceptron (MLP) based layer. 
        To guarantee the global descriptor $\hat{f}_{GSM,\uppsi}(\mathbf p_i^{(v)},\mathbf p_j^{(v)}-\mathbf p_i^{(v)})$ conditioning on $ f_{PIM}(\mathbf p_i^{(v)}, \mathbf Q_{\mathbf p_i^{(v)},K}) $, we need to construct the corresponding decoder and build relative constraint loss, which is shown in Section~\ref{subsec:CCD}.
        corresponding decoder and build relative constraint loss, which is shown in Section~\ref{subsec:CCD}.

    \subsection{Conditional Constrain Decoder}
    \label{subsec:CCD}
    
    As mentioned, the concatenation of GSM's and PIM's outputs can be viewed as Hierarchical Geometry Descriptor $F_{HGD}$ with dimension $N_v\times(C_{GSM}+J*K)$.
    After feeding $F_{HGD}$ into several per-point MLPs that corresponds to Recons MLPs in Fig.~\ref{fig:framework}, reconstructed point features are generated. 
    Our implementation takes two reconstruction MLPs with input/output dimensions as [576, 256] and [256,128]. 
    Denoting the output channel dimension of segmentation as $C_{out}$, the dimension of output point features $MLP_{Recons}(F_{HGD})$ is $C_{out} * 16$. 
    Then two branches are following, among which one is responsible for segmentation prediction, the other one provides adversarial learning for point identity descriptor.
     
    For the first branch, four Seg MLPs (Fig.~\ref{fig:framework}) with input/output dimensions $[C_{out} * 16, C_{out} * 8]\rightarrow[C_{out} * 8, C_{out} * 4]\rightarrow[C_{out} *4, C_{out} * 2]\rightarrow[C_{out} *2,C_{out}]$ take $MLP_{Recons}(F_{HGD})$ as the input and generate the segmentation results with dimension $C_{out}\times N_v$. The segmentation loss is a Cross-Entropy loss between the sparse annotation and the estimated labels for voxelized-level points. The loss is denoted as $\mathcal{L}_{Seg}$.
    After obtaining the sparse annotations, we perform radius nearest neighbors search to broadcast the annotations to all dense points as the post-processing step.
     

    The output of Point Identity Module, $f_{PIM}(\mathbf p_i^{(v)}, \mathbf Q_{\mathbf p_i^{(v)},K})$, can be viewed as the Point Identity Descriptor and denoted as $F_{PID}$. 
    Then in the second branch, the Point Identity Generator (Fig.~\ref{fig:framework}) is a MLP that takes $MLP_{Recons}(F_{HGD})$ as the input, and outputs the generated descriptor $\hat{f}_{PIM}(\mathbf p_i^{(v)}, \mathbf Q_{\mathbf p_i^{(v)},k})$.
    Denoting the generated Point Identity Descriptor as $\hat{F}_{PID}$, instead of directly designing a reconstruction loss between $F_{PID}$ and $\hat{F}_{PID}$, we construct a Generative Adversarial Networks (GAN) based loss metric, which is achieved by adding an additional discriminator loss on $F_{PID}$ difference $\delta_{PID} = F_{PID}-\hat{F}_{PID}$. 
    The motivation behind is to enhance the coupling between the global-local spatial connections descriptor  $\hat{f}_{GSM,\uppsi}(\mathbf p_i^{(v)},\mathbf p_j^{(v)}-\mathbf p_i^{(v)})$ and the local identity descriptor $f_{PIM}(\mathbf p_i^{(v)}, \mathbf Q_{\mathbf p_i^{(v)},K})$. We generate a random noise feature $\sigma_{PID}$ with the same dimension as $\delta_{PID}$ whose elements are random numbers ranging from 0 to $1e$-3. 
    The Point Identity Discriminator (Fig.~\ref{fig:framework}) views the random noise feature $\sigma_{PID}$ as the real data and $\delta_{PID}$ as the fake data.
    Denoting the Point Identity Discriminator as $D$, during the training process, the adversarial loss and discriminator loos can be formulated as:
        \begin{align}
            &\mathcal{L}_{Adv}=\log (1- D(\delta_{PID})) \label{loss_Adv} \\
            &\mathcal{L}_{Dis}=\log [D(\delta_{PID})]+ log[1-D(\sigma{PID})]
            \label{loss_Dis}
        \end{align}

   \begin{figure*}[ht]
    	\centering
        \includegraphics[width=0.7\linewidth]{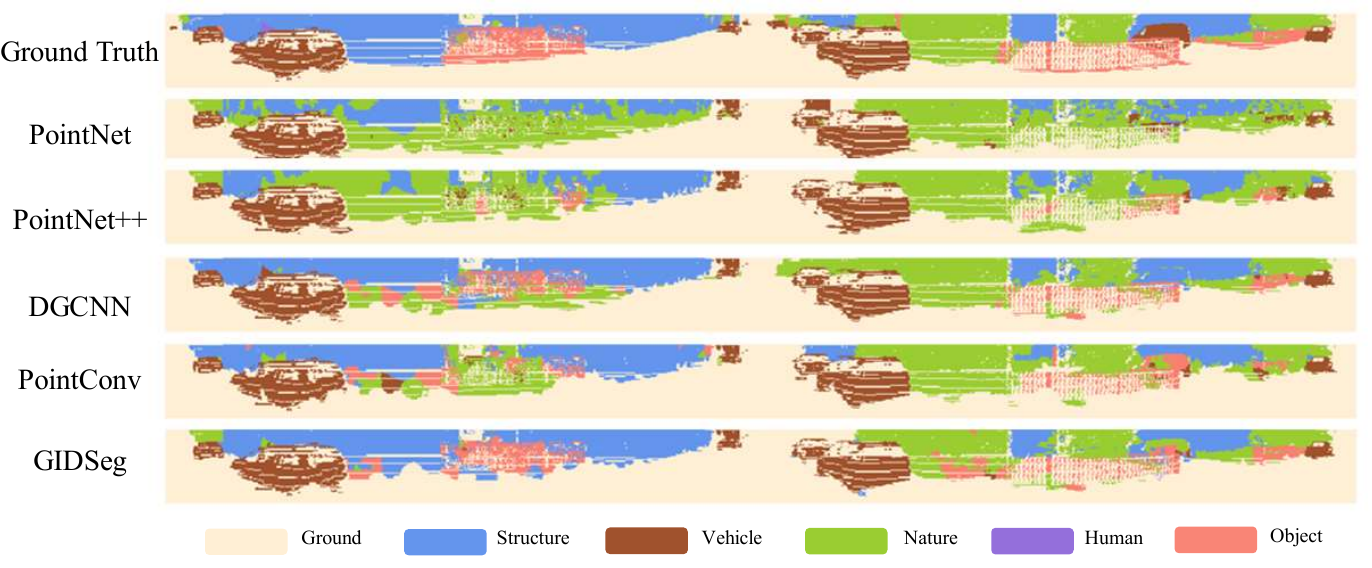}
    	\caption{\textbf{Segmentation results on \textit{SemanticKITTI} dataset.} Spherical view projection of 3D semantic segmentation results of different point-based methods.}
    	\label{fig:point-kitti}
    \end{figure*}
    
        \begin{figure*}[t]
    	\centering
        \includegraphics[width=0.7\linewidth]{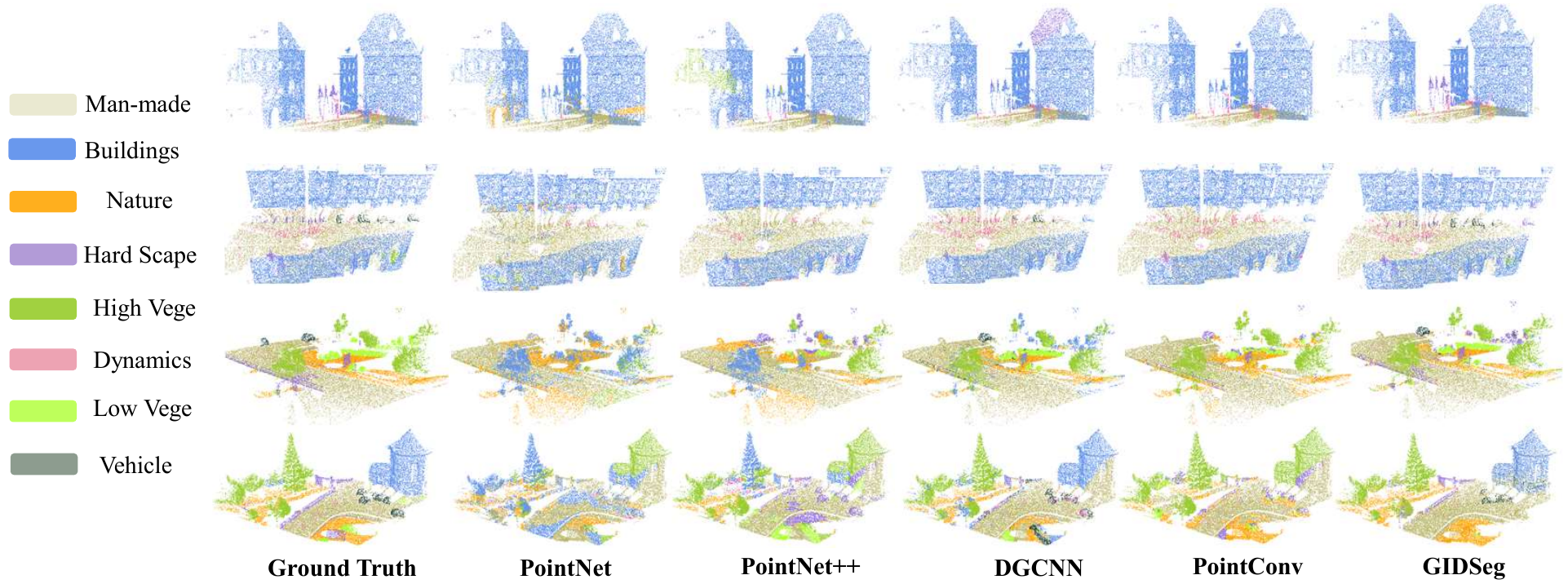}
    	\caption{\textbf{Semantic segmentation on \textit{Semantic3D}~\cite{data:semantic3D}.}
    	GIDSeg can achieve robust segmentation results comparing with other point-based methods, even for small objects.
    	}
    	\label{fig:point-semantic3d}
    \end{figure*}

\section{Experiment Setup}
\label{sec:EXP_setup}
    In this section, we evaluate the performance of GIDSeg and relative Point- and CNN- based 3D segmentation methods on \textit{Semantic3D}~\cite{data:semantic3D} and \textit{SemanticKITTI}~\cite{Data:semanticKitti} dataset. 
    All experiments are implemented with two Nvidia RTX 2080Ti GPU cards and $64$G RAM on an Ubuntu 18.04 system.

    \begin{table}[t]
        \centering
        \caption{Dataset splitting for different datasets.}
        \begin{tabular}{c c c c c}
        \toprule
         & Training & Validation & Inference \\ \midrule
        \textit{Semantic3D}~\cite{data:semantic3D}        & $18,007$ & $13,682$ & $16,458$ \\
        \textit{SemanticKITTI}~\cite{Data:semanticKitti}  & $19,000$ & $3,000$ & $9,005$ \\ 
        \bottomrule
        \label{table:dataset}
        \end{tabular}
    \end{table}

    \subsection{Dataset Overview}
    \label{subsec:dataset}
    
    Our experiment is performed on two datasets:
    \begin{itemize}
        \item \textbf{Semantic3D}~\cite{data:semantic3D}, which is recorded by a Surveying-grade laser scanner for large-scale outdoor segmentation task, includes 15 training scenes and eight classes labels (i.e, Man-made terrain, Natural terrain, High vegetation, Low vegetation, Buildings, Hard scape, Scanning artefacts and Cars).
        To generate training/inference point cloud samples, we uniformly extract sub-maps from the 15 training scenes, each sub-map covers an area of $40\times40\times10$ meters in 3D space of original map.
        \item \textbf{SemanticKITTI}~\cite{Data:semanticKitti} is the dense annotation of the \textit{KITTI}~\cite{data:kitti} odometry dataset, which include 28 classes labels and 21 odometry trajectories gathered around the mid-size city of Karlsruhe.
        In this work, we merged the 28 classes into 6 classes (Ground, Structure, Vehicle, Nature, Human and Object) as the official \textit{SemanticKITTI} suggested, namely Ground, Structure, Vehicle, Nature, Human and Object. 
        Each LiDAR scan contains around $120,000\sim 150,000$ points and can scan up to $60\times60\times10$ meters in 3D space.
    \end{itemize}
    
        As mentioned in Section.~\ref{sec:HGE}, GIDSeg requires both spares points and their corresponding radius kernelized features.
        In both \textit{Semantic3D} and \textit{SemanticKITTI} dataset, we down-sample the raw point cloud into $10,000$ points with a voxelization operation (voxel resolution is $0.2m$).
        We uniformly select $4,096$ points from the point cloud, and use sparse annotations ($1\%$, $5\%$, $10\%$,...,$100\%$ for each class within the point cloud) in the training procedure.

        There are several hyper-parameters need to be selected based on the data and object shapes 1) the voxel resolutions of both global and local points, 2) the radius of local points $R$, 3) the number of $K$ neighbors for global points and 4) the number of kernels $J$ to describe each point identity descriptor.
        In our setup, since both SemanticKITTI and Semantic3D are outdoor datasets, the dense map we extracted is a circle area with a radius of 30m. 
        The voxel resolution of local points is $0.1m$ and local radius $R=0.5m$ to generate enough neighbors ($>32$) for kernel feature extraction. 
        A sequence of four radius kernel function is used to encode the neighboring points into local features.
        The parameters within the radius kernel function is selected by gradient-decent searching to enhance feature description ability of radius kernelized features.
        To verify the suitable neighbors $K$ and kernel size $J$, we use different $[K,J]$ values and analyze their relative effects in Section~\ref{subsec:abstudy}.

         \begin{table*}[t]
	\caption{Segmentation accuracy on \textit{Semantic3D} and \textit{SemanticKITTI} datasets.}
	\begin{scriptsize}
	\renewcommand{\arraystretch}{1}
	\centering
	\begin{center}
	    \resizebox{\textwidth}{!}
	    {
        \begin{tabular}{ l c c c c c c c c c c c c}
        \toprule
        \multicolumn{12}{c}{{Semantic3D}} \\
        \toprule
        &  \makecell[c]{OA \\ (\%)} & \makecell[c]{mAcc \\ (\%)} & \makecell[c]{mIoU \\ (\%)} & \makecell[c]{man-made \\ terrain} & \makecell[c]{natural \\ terrain} & \makecell[c]{high \\ vegetation} & \makecell[c]{low \\ vegetation} & {buildings} & \makecell[c]{hard \\ scape} & \makecell[c]{scanning \\ artefacts} & {cars} \\
        \midrule
        PointNet\cite{3D:pointnet} & 63.9 & 31.6 & 22.4 & 56.6 & 26.1 & 22.9 & 1.6 & 61.7 & 0.7 & 9.6 & 0.0 \\
        PointNet$++$\cite{3D:pointnet2} & 71.2 & 43.0 & 31.6 & 62.5 & 45.8 & 33.6 & 4.2 & 71.8 & 12.6 & 22.0 & 0.0\\
        PointConv\cite{3D:PCNN} & 75.0 & 52.0 & 37.4 & 64.0 & 49.8 & 51.0 & 15.6 & 77.3 & 14.1 & 27.4 & 0.0\\
        DGCNN\cite{3D:DGCNN} & 85.0 & 65.6 & 53.2 & 77.6 & 74.0 & 70.8 & 30.0 & 84.2 & 26.1 & 37.8 & 25.3\\
        \midrule
        SqueezeSeg\cite{3D:sqseg} &  61.8 & 24.4 & 16.2 & 43.9 & 12.4 & 17.9 & 0.0 & 53.7 & 0.1 & 1.4 & 0.0\\
        SqueezeSegV2\cite{3D:sqseg2} & 83.7 & 40.3 & 33.0 & 66.0 & 22.5 & 40.1 & 9.7 & 77.4 & 25.2 & 14.6 & 8.2\\
        SqueezeSegV3\cite{3D:squeezesegv3} & \textHT[88.6] & 50.8 & 44.3 & \textHT[80.8] & 35.1 & 52.3 & 22.9 & \textHT[89.2] & 37.1 & 24.8 & 23.4\\
        Darknet21\cite{3D:rangenet++} & 86.5 & 46.9 & 39.7 & 68.9 & 25.8 & 45.3 & 21.3 & 80.3 & 35.5 & 19.3 & 21.2\\
        Darknet53\cite{3D:rangenet++} & 87.1 & 49.1 & 41.6 & 70.1 & 26.2 & 45.5 & 23.4 & 81.0 & 38.3 & 22.4 & 25.5\\
        \midrule
        GIDSeg & 87.9 & \textHT[68.6] & \textHT[61.2] & 79.6 & \textHT[78.9] & \textHT[77.7] & \textHT[40.2] & 88.5 & \textHT[41.5] & \textHT[42.6] & \textHT[40.6]\\
        \bottomrule
        \end{tabular}
        }
        \resizebox{\textwidth}{!}{%
        \begin{tabular}{ l c c c c c c c c c}
        \multicolumn{10}{c}{SemanticKITTI} \\
        \toprule
        & {OA(\%)} & {mAcc(\%)} & {mIoU(\%)} & {ground} & {structure} & {vehicle} & {nature} & {human} & {object} \\
        \midrule
        PointNet\cite{3D:pointnet} & 71.5 & 46.9 & 36.1 & 76.8 & 41.5 & 48.7 & 49.5 & 0.0 & 0.1\\
        PointNet$++$\cite{3D:pointnet2} & 78.9 & 53.9 & 44.2 & 79.6 & 56.6 & 64.5 & 61.0 & 0.0 & 3.5\\
        PointConv\cite{3D:PCNN} & 80.3 & 55.7 & 46.6 & 78.2 & 63.8 & 64.6 & 64.2 & 0.0 & 8.9 \\
        DGCNN\cite{3D:DGCNN} & 84.9 & 61.0 & 53.2 & 83.6 & 71.2 & 77.0 & 71.4 & 0.0 & 15.3\\
        \midrule
        SqueezeSeg\cite{3D:sqseg} & 71.5 & 47.2 & 34.8 & 83.7 & 43.6 & 40.8 & 40.5 & 0.0 & 0.1\\
        SqueezeSegV2\cite{3D:sqseg2} & 89.5 & 61.5 & 54.2 & 91.3 & 76.3 & 73.1 & 73.7 & 0.0 & 0.1\\
        SqueezeSegV3\cite{3D:squeezesegv3} & \textHT[94.8] & \textHT[70.1] & \textHT[64.0] & 95.2 & \textHT[83.5] & \textHT[84.4] & \textHT[83.9] & 1.2 & 10.8\\
        Darknet21\cite{3D:rangenet++} & 91.8 & 67.0 & 60.2 & 94.4 & 79.8 & 82.4 & 79.1 & 0.0 & 25.6\\
        Darknet53\cite{3D:rangenet++} & 93.2 & 69.3 & 63.0 & \textHT[95.3] & 82.6 & 83.9 & 82.4 & 0.9 & 33.8\\
        \midrule
        GIDSeg & 88.3 & 67.3 & 59.8 &  85.4 & 77.0 & 80.1 & 78.9 & \textHT[4.8] & \textHT[30.8]\\
        \bottomrule
        \end{tabular}
        }
	\end{center}
	\label{table:acc_point}
	\end{scriptsize}
    \end{table*}
    
    \subsection{Performance Analysis}
    \label{subsec: cmp_point}
    We compared our GIDSeg with four state-of-the-arts point-based methods: PointNet~\cite{3D:pointnet}, PointNet~\cite{3D:pointnet2}, DGCNN~\cite{3D:DGCNN} and PointConv~\cite{3D:pointconv}, and four 2D CNN-based methods: SqueezeSeg~\cite{3D:sqseg}, SqueezeSegV2~\cite{3D:sqseg2}, SqueezeSegV3~\cite{3D:squeezesegv3} and Darknet~\cite{3D:rangenet++}.
    All the point-based methods above are trained with the sparse point sets, which is $4,096$ points counting $5\%$ in the original point cloud.
    The only difference between our GIDSeg's inputs and the inputs of the other methods are the additional point-identity features as described in above section.
    All the CNN-based methods are trained with all the points, which are projected into sphere views.
    Table~\ref{table:acc_point} presents the quantitative results of different approaches on \textit{Semantic3D} and \textit{SemanticKITTI} datasets.
    Fig.~\ref{fig:point-kitti} and Fig.~\ref{fig:point-semantic3d} further shows the qualitative comparisons on the two datasets.
    

    Comparing with other point-based methods, the hierarchical feature encoding ability of our GIDSeg enhances the segmentation with only sparse annotations.
    As we can see in Table~\ref{table:acc_point}, we can notice that with $5\%$ annotations of the original point cloud, GIDSeg can achieve better segmentation performance than all other point-based methods on both datasets.
    GIDSeg leads the strongest baseline DGCNN~\cite{3D:DGCNN} by $3.9\%$ in overall accuracy on \textit{Semantic3D} dataset and $3.4\%$ on \textit{SemanticKITTI} dataset.
    This performance shows GIDSeg's generalization ability of doing inference with 3D segmentation with sparse annotations.
    Comparing with the CNN-based methods which are trained with fully points, GIDSeg can reach the similiar segmentation accuracy with only sparse annotations.
    The performance of CNN-based methods are different on two datasets, this is because the projections on \textit{Semantic3D} contain lots of overlaps than \textit{SemanticKITTI}.

    As shown in Table.~\ref{table:time}, we also analysis the mIoU under different annotation levels on the KITTI datasets.
    We also compared the state-of-the-art 3D segmentation method CylindrNet~\cite{3D:CylinderNet} and also the weakly supervised learning method~\cite{3D:weakly}. 
    And results show that under sparse annotations, our method outperform the others.
    While CylindrNet is better than our method under $100\%$ annotations, but it cannot achieve good performance with sparse annotations.
    Though the Weakly method is robust in indoor environment and shape datasets~\cite{3D:weakly}, the generalization ability in outdoor environments is limited.
    The average inference time of for GIDSeg is $114.7ms$, which include pre-processing time $110ms$ and online inference time $4.7ms$.
    
         \begin{table}[ht]
    	\caption[m1]{The mIoU under different annotation levels on the KITTI dataset.}
    	\begin{scriptsize}
    	\renewcommand{\arraystretch}{1}
    	\centering
    	\begin{center}
    		\begin{tabular}{  l |  c | c | c} 
    		    \toprule & mIoU ($1\%$)  & mIoU ($10\%$)  & mIoU ($100\%$)\\ 
                \hline
                PointNet$++$~\cite{3D:pointnet2}    & 37.9\% & 48.4\% & 51.4\% \\
                DGCNN~\cite{3D:DGCNN}               & 41.7\% & 54.8\% & 58.1\% \\
                CylindrNet~\cite{3D:CylinderNet}    & 38.3\% & 49.9\% & 64.5\% \\
                Weakly~\cite{3D:weakly}             & 39.4\% & 55.2\% & 58.32\% \\
                \hline
                GIDSeg                              & 45.8\% & 60.1\% & 63.8\% \\
                \bottomrule
    		\end{tabular}
    	\end{center}
    	\label{table:time}
    	\end{scriptsize}
        \end{table}

        \begin{table*}[t]
    	\caption{Semantic segmentation results on \textit{Semantic3D} and \textit{SemanticKITTI} dataset with GIDSeg's different PIM configurations.
        }
    	\renewcommand{\arraystretch}{1}
    	\centering
    	\begin{center}
    	    \resizebox{\textwidth}{!}{%
    		\begin{tabular}{  l | c  c  c | c  c  c} 
    		    \toprule
    		    & \multicolumn{3}{c}{Semantic3D} & \multicolumn{3}{c}{SemanticKITTI} \\
    		    \hline
                Network Configurations & mIoU(\%) & OA(\%) & mAcc(\%) & mIoU(\%) & OA(\%) & mAcc(\%)\\ 
                \midrule
                Without PIM module & 53.9 & 84.4 & 66.3 & 54.8 & 85.4 & 61.9\\
                PIM + bell-shaped curve ([J=1, K=8]) & 54.7 & 84.3 & 66.9 & 55.1 & 85.9 & 62.5\\
                PIM + bell-shaped curve ([J=1, K=16]) & 55.6 & 85.1 & 68.9 & 56.3 & 87.3 & 63.2\\
                PIM + bell-shaped curve ([J=1, K=32]) & 55.8 & 84.5 & 68.6 & 56.5 & 86.6 & 63.3\\
                PIM + bell-shaped curve ([J=4, K=16]) & 57.2 & 84.7 & 70.4 & 57.9 & 87.6 & 65.4\\
                PIM + bell-shaped curve ([J=4, K=16]) + Dis &  \textbf{61.2} & \textbf{87.9} & \textbf{73.5} & \textbf{59.8} & \textbf{88.3} & \textbf{67.3}\\
                \bottomrule
    		\end{tabular}
    		}
    	\end{center}
    	\label{table:ablation}
        \end{table*}

    \begin{figure*}[t]
    	\centering
        \includegraphics[width=0.7\linewidth]{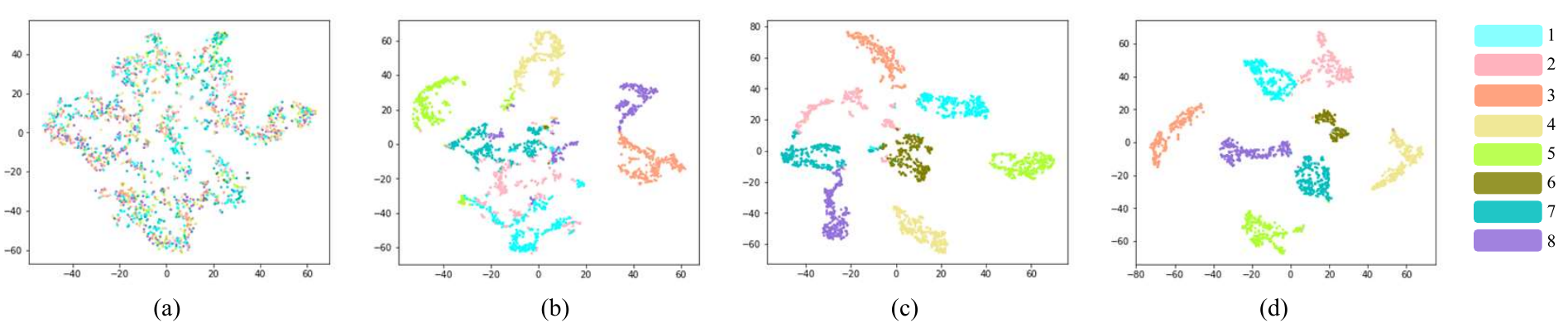}
    	\caption{\textbf{t-SNE visualization.} 
    	 Feature distribution on Semantic3D datasets with 8 classes (as shown with different colors): (a) input 3D points, (b) output features without PIM (c) output features without Discriminator (d) output features with all modules.
        }
    	\label{fig:tSNE}
    \end{figure*}

    \subsection{Ablation Study}
    \label{subsec:abstudy}
        To analyze the impact of points' sparsity level on the segmentation accuracy, we make comparison between our method and CNN-based methods.
        We conducted these experiments on the \textit{SemanticKITTI} datasets.
        GIDSeg is trained with different percentages of the raw data (range from $1.25\%$ to $15\%$).
        We also conducted ablation studies on both \textit{Semantic3D} and \textit{SemanticKITTI} datasets to further investigate in the effects of different configurations of GIDSeg's PIM module.
        We tested the segmentation performance by adding/removing PIM module, utilizing different $K$ values in PIM module, and testing different $M$ in Edge Conv layers.
        A careful selection of bell-curve shapes is performed to ensure that multiple levels of geometric features are captured.
        We also verify the segmentation performance in the PIM module with/without the discriminator, which we claim to be essential for conditional decoding as described in Section.~\ref{subsec:CCD}.
        
           \begin{figure}[h]
        	\centering
            \includegraphics[width=\linewidth]{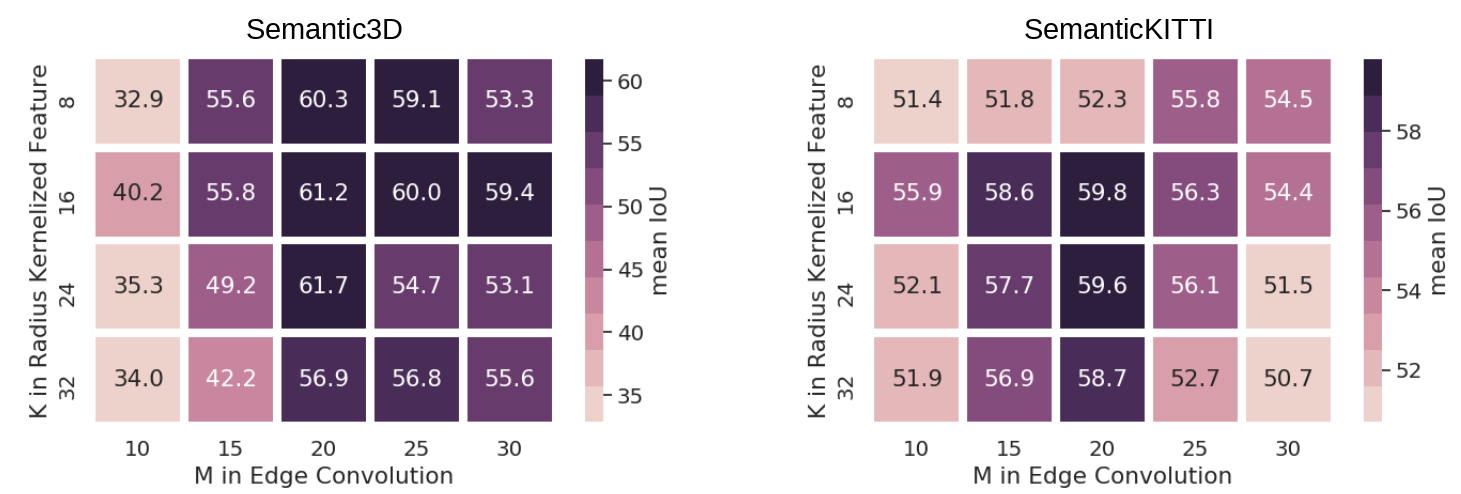}
        	\caption{\textbf{Ablation study for the radius kernel and edge convolution.} The mean IoU under different settings of $K$ of radius kernel function and $M$ in edge convolution layer on (a) Semantic3D, (b) SemanticKITTI dataset.}
        	\label{fig:ablation_km}
        \end{figure}
    
        As shown in Table.~\ref{table:ablation}, the case that without PIM module (i.e., only use sparse voxelized points, without point-wise radius kernel features) is worse than all other cases with PIM module.
        In the later cases, suitable radius kernelized feature can improve the segmentation performance, and the discriminator of PIM can further enhance the segmentation accuracy by adding conditional constraints.
        If the number of neighbors $K$ is too small, the neighbors of the voxelized points cannot cover enough information, preventing local features from being effectively captured. 
        While if $L$ is so large that points belonging to different classes are included, local features cannot be represented precisely. 
        Fig.~\ref{fig:ablation_km} shows the relationship between edge convolution and radius kernelized feature.
        As we can see the best parameters for \textit{Semantic3D} datasets are [K=24, M=20], while for \textit{SemanticKITTI} are [K=16, M=20].
        One limitation of current work is that GIDSeg cannot select the best $K$ and $M$ for point cloud datasets, and such hyper-parameters will vary based on the points distribution.
        In further work, we can investigate apply gradient decent approach to automatically select the best parameters for new point cloud datasets.
        
        To fully understand the performance of the proposed GIDSeg method, we utilize the t-Distributed Stochastic Neighbor Embedding (t-SNE)~\cite{tool:tSNE} algorithm to visualize the segmentation distributions, which calculates a similarity measurement between pairs of instances in the high dimensional space. 
        In Fig.~\ref{fig:tSNE}, we analysis the feature distribution of both original 3D points and the second to last layer of Seg MLPs module in Fig.~\ref{fig:method}.
        We randomly sample $300$ points from the full area for each category and plot the feature distribution as illustrated.
        Without adversarial learning, the learned point identity module can have limited generalization ability, and easily mixture with other patterns.
        From both Table.~\ref{table:ablation} and Fig.~\ref{fig:tSNE}, we can notice that, with the adversarial learning, PIM can enhance enhance the generalization ability of the GIDSeg by parallelly updating the Point Identity Module, Point Identity Generator Module and Recons MLPs module.

    \subsection{Discussion}
\begin{figure}[h]
        	\centering
            \includegraphics[width=0.8\linewidth]{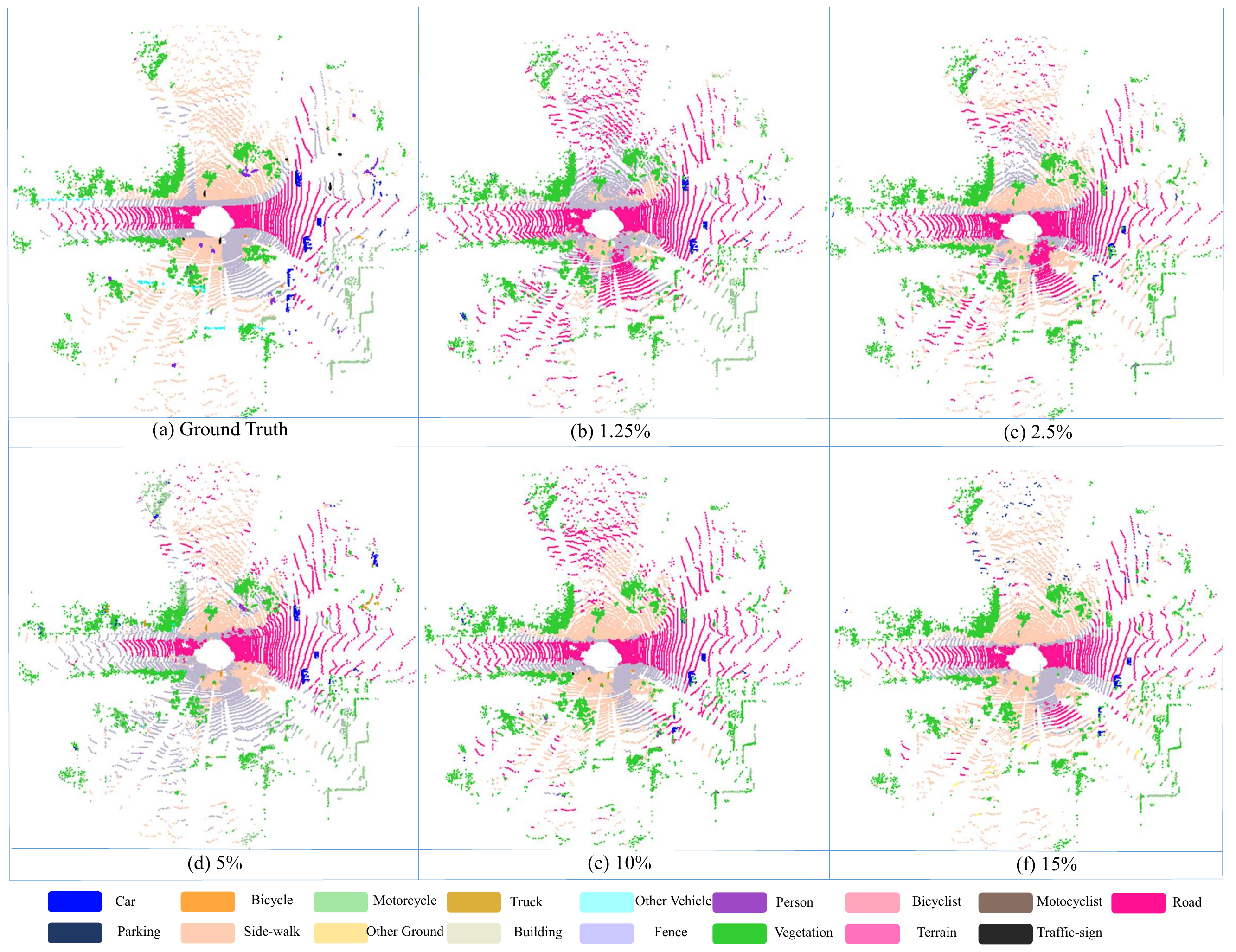}
        	\caption{\textbf{Segmentation results on \textit{SemanticKITTI} dataset using different proportion of points with 19 classes.}  
            }
        	\label{fig:part2}
        \end{figure}

         The complexity of our method is related to the semantic classes. 
         GIDSeg is mainly suitable for large objects (cars, trees, buildings, etc), this is due to the fact that our semantic extraction is based on the combination of high resolution geometry features and voxel level geometry structures, which makes are method less sensitive to small objects.
         As we can see in the Fig.~\ref{fig:part2}, we also trained the GIDSeg for the 19 class on SemanticKITTI, while some class with smalls shapes will be easily ignored. 
         Another limitation of our method is in the pre-processing step, where KNN search and kernel feature extraction are time-consuming.
         In the future work, we will apply parallel processing to speed up the local feature extraction.

\section{Conclusion}
In this paper, we introduced an end-to-end learning method, GIDSeg, that learns 3D semantic segmentation with only sparse labels. 
It offers an accurate segmentation from only $5\%$ annotations of the original point cloud by capturing the hierarchical geometry features from both voxel-level global and dense-level local geometry structures.
The final segmentation results are predicted based on the joint feature distribution with the assistance of our conditional constraints decoder module.
Experimental results on two challenging datasets demonstrate the effectiveness and generality of our method.
With only sparse annotations, GIDSeg achieves the state-of-the-art on point-based methods, and can also surpass fully-annotated CNN-based methods.
The limited requirement for computation resource and efficient real-time inferencing make our method possible to imply on mobile robot platform.

\bibliographystyle{IEEEtranTIE}
\bibliography{Refer}\ 

\end{document}